# Improving Reliability and Explainability of Medical Question Answering through Atomic Fact Checking in Retrieval-Augmented LLMs

Juraj Vladika[†1], Annika Domres[†2], Mai Nguyen[2], Rebecca Moser[2], Jana Nano[2], Felix Busch[3], Lisa C. Adams[3], Keno K. Bressem[3,4], Denise Bernhardt[2], Stephanie E. Combs[2,5,6], Kai J. Borm[2], Florian Matthes[1], Jan C. Peeken*[2,5,6]

1 Department of Computer Science; TUM School of Computation Information and Technology; Technical University of Munich, Garching, Germany
2 Department of Radiation Oncology, TUM University Hospital Rechts der Isar, TUM School of Medicine and Health, Technical University of Munich, Munich, Germany
3 Department of Diagnostic and Interventional Radiology, TUM University Hospital Rechts der Isar, TUM School of Medicine and Health, Technical University of Munich, Munich, Germany
4 Institute for Cardiovascular Radiology and Nuclear Medicine, TUM University Hospital, German Heart Center Munich, TUM School of Medicine and Health, Technical University of Munich, Munich, Germany
5 Institute of Radiation Medicine (IRM), Helmholtz Zentrum München (HMGU)
6 German Consortium for Translational Cancer Research (DKTK), Partner Site Munich, Munich, Germany

† These authors contributed equally to this work.

***Corresponding Author**: Jan C Peeken, MD PhD
Email: jan.peeken@tum.de
Address: Department of Radiation Oncology, TUM School of Medicine and Health, TUM University Hospital Rechts der Isar, Munich, Germany

## Abstract:

Large language models (LLMs) exhibit extensive medical knowledge but are prone to hallucinations and have low fact-level explainability, limiting clinical adoption and regulatory compliance. Existing approaches, such as Retrieval Augmented Generation, partially address these issues by grounding answers in source documents, however, the aforementioned problems persist. We propose a novel atomic fact-checking framework, designed to enhance the reliability and explainability of LLMs in medical long-form question answering. By decomposing generated answers into discrete atomic facts, and verifying each against an authoritative knowledge base of medical guidelines, precise identification and correction of incorrect statements, alongside explicit linkage to supporting literature, are enabled. Extensive evaluation, including multi-reader assessments by medical experts and an automated open Q&A benchmark, demonstrated significant improvements in factual accuracy, achieving a 40% improvement in overall answer quality. This work addresses key prerequisites towards a more trustworthy and reliable clinical application of LLMs.

## Introduction:

Large language models (LLMs) exhibit extensive medical knowledge.[1] However, LLMs are prone to hallucinations that may lead to harmful medical advice, and their often inaccurate citations reduce overall explainability.[2] Such limitations also complicate medical product certifications.[3] Current methods such as Retrieval Augmented Generation (RAG) partially address these issues by grounding answers in source documents.[4]

In RAG, the most common setup is to divide source texts into discrete chunks, embed into a vector space, and retrieve as needed to ground LLM responses in updatable, authoritative information. Prior research shows that RAG can improve answer quality in medical Q&A.[5] Nevertheless, answers can still contain *hallucinations* – statements that are factually incorrect and contradict the established knowledge. Additionally, fact-by-fact explainability of answers remains low, especially for complex medical queries. Existing approaches do not address the need for validating, backtracing and correcting each individual claim within a long-form response.

Methods to increase the factuality of LLM outputs can involve additional pre-training or fine-tuning of models, which are computationally expensive and impossible for closed-source models. Hence, post-hoc methods are emerging, where LLMs self-correct their responses only after they are generated.[6] A promising method is automated fact-checking, which detects individual facts from the generated response that contain information contradicting the authoritative knowledge and rewrites these facts and final responses using correct information. Originating from journalism where it is performed manually, fact-checking is increasingly used for hallucination correction.[7] However, existing approaches mostly focus on the encyclopaedic and news domains and are underexplored for medical applications.[8]

We define an atomic fact as the smallest self-contained and verifiable unit of information in a response generated by an LLM[9] (e.g., "Trastuzumab is indicated for HER2-positive breast cancer"). We developed a framework that decomposes responses into atomic facts, each of which is independently verified against an authoritative vector database. This approach enables targeted correction of errors and direct tracing to source literature, thereby improving the factual accuracy and explainability of medical Q&A. To the best of our knowledge, this is the first application of such an "atomic fact-checking" approach to the medical domain.

We extensively evaluated the framework using multi-reader assessments by medical experts on three Q&A datasets we constructed, as well as using an automated open Q&A benchmark, demonstrating substantial improvements in factuality and explainability over baseline RAG-based LLM systems.

## Results:

Across all evaluation sets, answers were split into a median number of 7 atomic facts (interquartile range: 5–8.75), yielding a total number of 428, 404, 519 facts for the validation (50 Q&A), testing (60 Q&A), and tumorboard (40 Q&A) test sets, respectively. Evaluation on the validation set revealed the best architectural strategy as four shot prompting for answer generation, verdicting, and fact rewriting (ablation study: Supplementary Table S7, Figure S1). In the atomic fact correction step, retrieving new chunks for each fact instead of using the initial chunks (used for Q&A) for fact-checking and rewriting increased the overall performance (both precision and sensitivity).

Looping through all facts labelled as "FALSE" further increased the true positive rate, and positive predictive value while reducing the falsification of rewritten facts. The average false positive rate over three evaluation sets reduced from 2% to 1% to 0% throughout 3 iterations of the correction loop. Therefore, three iterations were chosen as maximum since that was enough in experiments to correct all unsupported facts. Ensembling for atomic veracity prediction and changes in temperature during ensembling did not yield a significant gain in performance (Supplementary Figures S2, S3). While it slightly increased the overall F1 score, it also reduced sensitivity, the most important metric in our system. Therefore, our final pipeline uses four-shot examples and three-step looping but no veracity prediction ensembling.

Adding few-shot examples to LLM prompts decreased the precision score (95, 71, 61 for zero, one, and four-shot) but increased the sensitivity (recall) score (44, 62, 78). Sensitivity was preferred because it is more important in our system to detect any hallucinations, but any false positives could be dismissed during the rewriting phase.

In the validation and test Q&A sets, this final framework achieved balanced accuracy scores of 87 % and 74%, with a hallucination detection of 50 % and 38 %, and an inaccuracy detection of 50% and 25%, respectively (Table 1). In the more complex tumor board test set, 25% of hallucinations and 24% of inaccuracies were found with balanced accuracies of 78%. The positive predictive value for false fact identification was 66%, 100%, and 97% for the validation, test, and tumor board set, respectively.

Overall answer improvements were seen in 20%, 12%, 40% of cases in the validation, test, and tumor board Q&A sets (Figure 2). Overall answer quality decreased in 8%, 0%, and 6 % of cases. The quality of the remaining answers remained the same.

Achieving results comparable to our own test set on the Neurology Q&A shows that overfitting was avoided. Presenting with a sensitivity of 52%, precision of 86% and balanced accuracy of 75.9%, the overall answer quality was improved, remained stable, worsened in 10%, 10% and 4%, respectively.

The AMEGA auto-evaluation analysis revealed a significant improvement in answer quality by applying our fact checking algorithm in contrast to RAG only for all LLMs tested ($p<0.001$ for 8/11 models, $p<0.01$ for 2/11, $p<0.05$ for 1/11) (Figure 3A, Supplementary Table S6). The overall best answer quality (27.3) among non-reasoning models was achieved by GPT4o-mini and Mistral 24B, after fact-checking has been applied. The largest improvement with fact-checking was seen for Llama 3.2 3B. A significant negative correlation was found between the logarithm of the LLM model parameter size (in billions) and model performance (*Pearson Correlation: -0.754, p-value: 0.031)*. Even for the reasoning model OpenAI-o1, the performance was significantly increased ($p<0.001$) while achieving the overall best model performance of 31.52.

An important aspect of the fact-checking framework is the explainability achieved by tracing each atomic fact to the most relevant chunk (passage) in medical guidelines. To assess this, we used the test Q&A set and compared three similarity definitions of the best-fitting chunks selected from the vector database. A simple Chain-of-Thought (CoT) prompt using GPT4o identified the correct chunk in 75% as the first choice and in 91.9% among the top three chunks, outperforming a complex CoT prompt (instructing the LLM to assign scores and then rank) and cosine similarity with a text-embedding model (Supplementary Figure S4, Supplementary Table S3 for prompts).

In the comparison of open-source medical and non-medical models (Table 2), MedGemma 27B performed best overall, with a balanced accuracy of 90%, sensitivity of 83%, and total improved answers of 30%, outperforming even the GPT-4o baseline. The high sensitivity score of MedGemma 27B was better compared to its Gemma 3 27B counterpart and to GPT-4o (83% vs. 46% vs. 78%).

The two tested Qwen models showed comparable sensitivity (55% vs. 56%), while Llama3 70B considerably outperformed OpenBioLLM 70B (61% vs. 33%). The same tendency holds true for their F1 scores and balanced accuracies. The positive predictive value was the highest for OpenBioLLM 70B, however, it had a markedly reduced sensitivity and elevated false negative rate.

Looking at the improved overall answer quality as the main fact-checking outcome, MedGemma 27B demonstrated superior performance compared to its generalist counterpart Gemma 3 27B, as well as GPT-4o baseline (30% vs. 20% vs. 20%) and all other evaluated models. The Qwen models showed moderate results (generalist 18% and medical 24%), while Llama 3 surpassed OpenBioLLM (8% vs. 4%). Among non-medical models, Llama3 achieved the highest balanced accuracy (79%).

Results of LLM-as-a-judge automated rubric evaluation are shown between the baseline, GPT-4o, and the best performing open-source model based on F1 score, MedGemma 27B, in Table 3. Results of all open-source models are available in Supplementary Table S9. Even though MedGemma 27B outperformed GPT-4o on the fact-checking task (atomic claim veracity prediction), GPT scored better in four rubrics, MedGemma in one rubric, while scores were almost equal in two rubrics. While MedGemma improved more answers in total (30% vs. 20%), certain aspects of its answers are still lacking compared to GPT. Most of MedGemma's answers had a significantly higher rubric score after the fact-checking process was performed. When evaluating all LLMs together, context faithfulness was significantly improved (Supplementary Table S9).

Comparison to the competing self-correcting framework Self-Refine[10] on AMEGA is shown in Table 4. The improvement rate using Self-Refine is better on more powerful models (GPT) while the smaller and less capable models (Gemma and Llama) benefit more from our atomic fact-checking framework, showing that smaller LLMs struggle with large feedback prompts but excel when tasks are decomposed into many shorter calls.

| Numbers in percent | Validation Set | Test Set | Tumor Board | NeurologyQA[2] |
|---|---|---|---|---|
| Sensitivity (Recall) | 78 | 47 | 44 | 52 |
| Specificity | 96 | 100 | 100 | 100 |
| Precision (PPV) | 66 | 100 | 97 | 86 |
| F1-score | 71 | 64 | 60 | 65 |
| Balanced Accuracy | 87.3 | 73.7 | 91.0 | 75.9 |
| TP improved atoms | 100 | 100 | 100 | 92 |
| FP falsified atoms | 0 | 0 | 0 | 0 |
| Hallucination rate | 1 | 2 | 1 | 0 |
| Hallucination detection | 50 | 38 | 25 | 0 |
| Inaccuracy rate | 3 | 1 | 5 | 1 |
| Inaccuracy detection | 58 | 50 | 24 | 50 |

**Abbreviations: TP: True Positives, FP: False Positives, PPV: Positive predictive value**

**Table 1: Human Evaluation Results on Four Benchmark Datasets.** We present results of different metrics as rated by medical expert annotators over our three constructed and one external Q&A datasets.

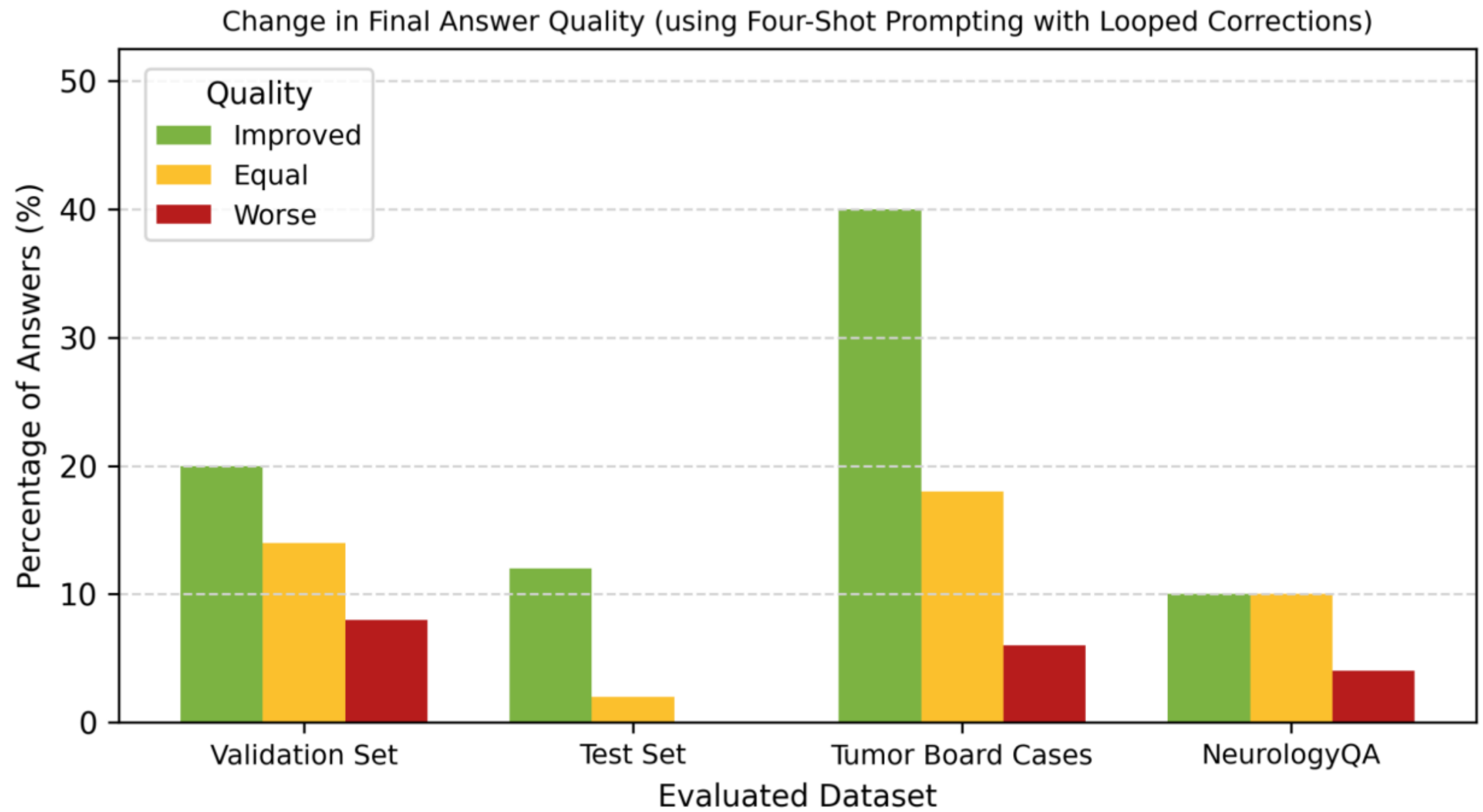

**Figure 2: Change in Final Answer Quality.** We show how the overall answer quality changed by comparing the initial vs. the final fact-checked answer. This is shown for our four evaluation Q&A datasets in three directions (improved, equal, worse), as rated by human expert annotators.

**A)**

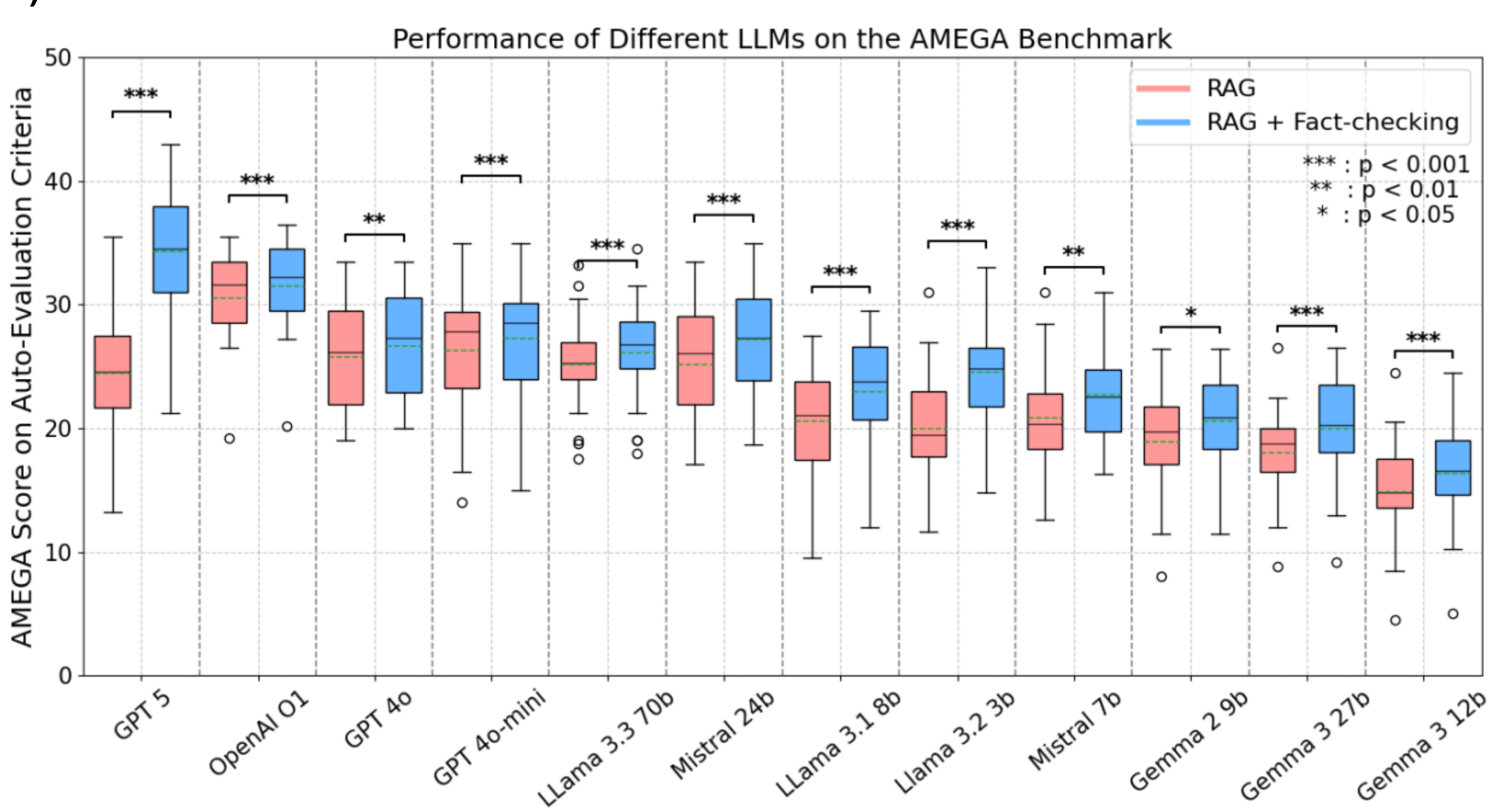

**B)**

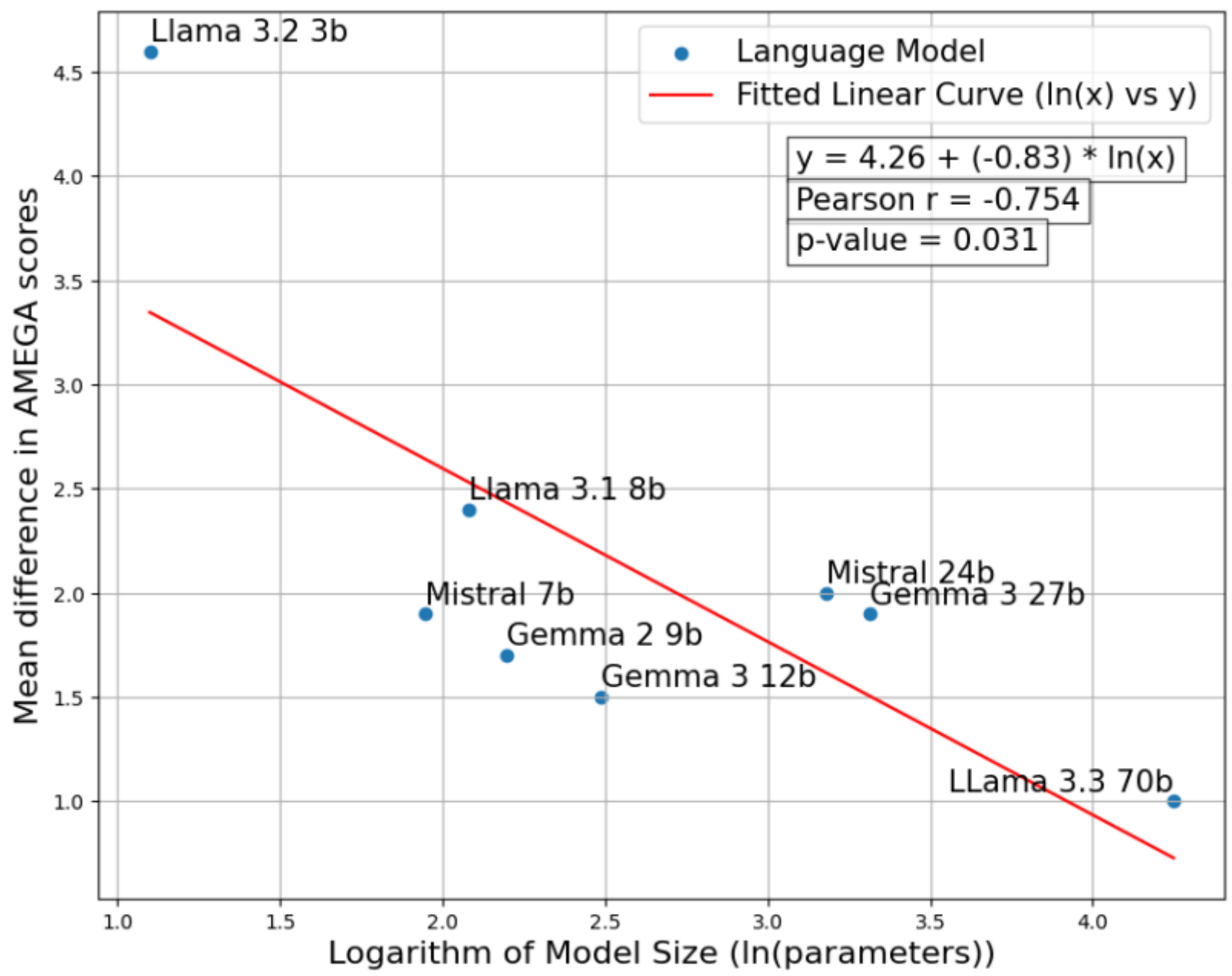

**Figure 3: Results of Auto-Evaluation on the AMEGA Benchmark. (A)** Scores for 12 different open- and closed-source LLMs, without and with answer fact-checking. **(B)** Fitted linear curve of the natural logarithm of LLM parameter size in billions vs. the average absolute score improvement of using fact-checking for that model.

| Numbers in percents | GPT4o (baseline) | Gemma3 27B | Med Gemma 27B | Llama3 70B | OpenBio LLM 70B | Qwen3 32B | Qwen3 Medical 32B |
|---|---|---|---|---|---|---|---|
| TP | 6 | 6 | 11 | 12 | 7 | 8 | 8 |
| FP | 3 | 3 | 3 | 2 | 1 | 2 | 3 |
| TN | 89 | 84 | 84 | 78 | 79 | 83 | 82 |
| FN | 2 | 7 | 2 | 8 | 14 | 6 | 7 |
| Sensitivity (Recall) | 78 | 46 | 83 | 61 | 33 | 56 | 55 |
| Specificity | 96 | 97 | 97 | 97 | 99 | 98 | 96 |
| Precision (PPV) | 66 | 68 | 79 | 83 | 88 | 80 | 72 |
| F1-Score | 71 | 55 | 81 | 70 | 48 | 66 | 63 |
| Accuracy | 95 | 90 | 95 | 90 | 86 | 92 | 90 |
| Balanced Accuracy | 87.3 | 71 | 90 | 79 | 66 | 77 | 76 |
| **TP improved atoms** | 100 | 100 | 91 | 98 | 100 | 82 | 89 |
| **FP falsified atoms** | 0 | 30 | 44 | 50 | 0 | 0 | 14 |
| **Overall answer quality improved** | 20 | 20 | 30 | 8 | 4 | 18 | 24 |
| **Overall answer quality equal** | 14 | 8 | 6 | 4 | 4 | 8 | 22 |
| **Overall answer quality worse** | 8 | 4 | 6 | 6 | 10 | 16 | 10 |
| **Hallucination rate** | 1 | 2 | 3 | 6 | 11 | 4 | 2 |
| **Hallucination detection** | 50 | 71 | 80 | 52 | 50 | 69 | 60 |
| **Inaccuracy rate** | 3 | 2 | 3 | 5 | 3 | 3 | 4 |
| **Inaccuracy detection** | 58 | 38 | 90 | 68 | 22 | 21 | 44 |

**Abbreviations: TP: True Positives, FP: False Positives, TN: True Negatives, FN: False Negatives, PPV: Positive predictive value**

**Table 2: Comparing the fact-checking performance of multiple open-source LLMs and the proprietary GPT-4o (baseline, used in previous experiments).** Focus was put on general-purpose vs. medical fine-tuned models of different sizes: Gemma 3 27B vs. MedGemma 27B, Llama 3 70B vs. OpenBioLLM 70B, Qwen 3 32B vs. Qwen 3 Medical 32B

| *LLM/Metric* | Correctness | | Complete-ness | | Clarity | | Context Faithfulness | | Coherence | | Medical Harmfulness | | Calibration | |
|---|---|---|---|---|---|---|---|---|---|---|---|---|---|---|
| *Baseline/Final* | B | F | B | F | B | F | B | F | B | F | B | F | B | F |
| GPT-4o | 66 (60–72) | 66 (60–73) | 93 (91–95) | 92 (90–94) | 87 (86–88) | 86 (85–87) | 88 (85–91) | 88 (84–91) | 89 (89–90) | 89 (89–90) | 95 (92–98) | 94 (91–97) | 87 (86–89) | 87 (85–89) |
| MedGemma 27B | 57 (49–65) | 58 (50–65) | 81 (76–87) | 86* (82–91) | 80 (78–82) | 83* (81–85) | 90 (87–92) | 92* (91–93) | 76 (71–81) | 80 (76–83) | 94 (92–95) | 95* (93–97) | 86 (84–87) | 87* (85–89) |

**Table 3: Comparison of LLM-as-a-judge rubric scores.** We show the final auto-evaluation scores of seven rubrics between the baseline model, GPT-4o, and the best-performing open-source model, MedGemma 27B. The scores are averaged across 50 answers on the validation set. Baseline (B) scores are for initial answers based on RAG. Final (F) scores are for final answers, after fact-checking was performed.
(* : $p < 0.05$) (95%-confidence intervals given below the averages)

| Model | Initial Score on AMEGA | with Fact-Checking (our approach) | with Self-Refine[10] (as baseline) |
|---|---|---|---|
| **GPT-4o** | 25.4 | 26.5 (+1.1) | 28.3 **(+2.9)** |
| **GPT-4o-mini** | 26.3 | 27.3 (+1.0) | 28.0 **(+1.7)** |
| **Gemma 3 27B** | 18.1 | 20.0 **(+1.9)** | 19.4 (+1.3) |
| **MedGemma 27B** | 21.8 | 24.2 **(+2.4)** | 22.8 (+1.0) |
| **Llama 3.2 3B** | 20.2 | 24.5 **(+4.3)** | 21.0 (+0.8) |

**Table 4: Comparison of our approach to the baseline on AMEGA.** We show the final auto-evaluation scores on AMEGA with five representative LLMs based on (1) initial responses; (2) responses corrected with our fact-checking framework; (3) responses corrected with the competing Self-Refine approach.

## Discussion:

Our study introduces a novel atomic fact-checking framework designed to enhance the reliability and explainability of LLMs used in medical Q&A. By decomposing LLM-generated responses into discrete atomic facts and rigorously verifying each against an authoritative vector database, the framework significantly reduced hallucinations and inaccuracies. Medical expert assessment and automated benchmarks demonstrated notable improvements in factual accuracy, achieving up to 40% overall answer improvement and 80% hallucination detection rate. Additionally, the framework achieved high explainability by tracing each atomic fact back to the most relevant chunks from the database, providing a granular, transparent explanation of the generated responses. Rubric-based autoevaluation revealed a LLM-dependency in score improvement, however, with a significantly increased source faithfulness across all LLMs.

Our framework increased the factual accuracy and overall quality of LLM-generated responses. Numerical hallucinations, such as incorrect drug dosages, were frequently identified and corrected, as were entity hallucinations, such as the conflation of different treatment procedures. In a small proportion of cases (8%, 0%, 6% and 4% across Q&A sets), the answer quality declined due to the retrieval of less relevant chunks or potential hallucinations were introduced during the correction process. However, since the proportion of answers that improved (20%, 12%, 40% and 10%) was substantially higher, the overall effect of fact-checking clearly enhanced the answer quality. Different degrees of improvement come down to varying complexity of questions across datasets and to different rates of atomic facts being identified as false and corrected.

Notably, the observed gain was strongest in real tumor-board questions. The most challenging dataset reached the highest rate of answer improvement (40%). The other three Q&A datasets, where questions were usually answerable directly from guideline passages, resulted with moderate improvement after fact-checking (20%, 12% and 10%). This suggests that the system contributes most when queries are complex and multi-factual, e.g. in clinical settings.

Analyzing the performance of medical fine-tuned LLMs in detecting incorrect facts, the model size appeared to be the most influential factor. Medical fine-tuning mostly rather worsened the performance, with MedGemma 27B marking an exception.
There are differences in fine-tuning of different models, hence performance differences are to be expected to some extent.

MedGemma 27B achieved notable results and improvement by fact-checking, outperforming even the GPT-4o baseline. Being a reasoning model (outputs “thinking” tokens before final response) likely helps its performance. The reasoning models have the additional benefit of higher interpretability of their answers. However, as measured by the LLM-as-a-judge rubrics, the quality of MedGemma 27B's answers was still lower than those of much larger closed-source GPT-4o, leaving space for future improvements.

RAG-based medical chatbots may qualify as Class IIa medical devices according to the European Medical Device Regulation.[3] Trustworthiness, the combination of explainability, traceability, and transparency, is a key prerequisite required under the MDR.[11] Although LLMs inherently explain their responses, these justifications can be misleading. Our framework provides backtracking to its source, achieving improved explainability, traceability, and transparency on fact level. Importantly, this process is outsourced from the internal reasoning of LLMs and directed towards a potential user.
We could demonstrate that the benefit of our framework was greater in smaller LLMs. Smaller LLMs had higher improvement using our fact-checking framework in benchmarks, including AMEGA, where our approach also outperformed the competing single-pass correction baseline. The even higher usability of the fact-checking algorithm in smaller LLMs is especially interesting, considering scenarios of potential on-premises deployment in medical institutions.

Our work has several limitations. As we sought to evaluate open Q&A capabilities, we designed novel evaluation datasets. As this evaluation process is time-consuming, we had to restrict our Q&A to 175 Q&As in total. These Q&As were based on information present in the provided guideline documents. We therefore included real anonymized patient cases from tumor boards as an even more realistic application. Furthermore, all the pipeline steps rely on LLM generation. It is possible that errors can propagate from one step to another. Future work could explore approaches to make the process more rigorous, such as Graph-RAG techniques of grounding data into graph structures for more robust generation. Finally, the entire fact-checking and rewriting process uses around 10 times more tokens than just the initial response generation (shown in Supplementary Table S11), which increases the costs and the latency of the system.

To conclude, we present a novel atomic fact-checking algorithm that identifies fact inaccuracies and hallucinations. Correcting these findings improves the overall answer quality while achieving fact-wise explainability, paving the way for more trustworthy clinical use of LLMs.

## Methods:

### Atomic Fact-checking Framework

The LLM-TRIPOD checklist is provided in the Supplement. Our fact-checking framework consists of five steps, as shown in Figure 1: (1) generate an initial RAG-based response to the question; (2) split the response into atomic facts; (3) determine the veracity of each fact (categories: “TRUE” and “FALSE”) based on newly retrieved chunks; (4) rewrite the facts detected as incorrect and loop through steps 3 and 4 until all facts are correct or for a maximum of three iterations; (5) rewrite the full response by incorporating the new rewritten facts.

All five steps are implemented with dedicated LLM prompts, provided in Table S1 in the *Supplementary material*. For steps 1-4, in-context learning (ICL) with four expert-annotated examples per prompt is used to enforce consistent atomic fact extraction and verdicting. All prompts and examples were created by a medical expert (JCP, 9 years of experience in Radiation Oncology). Prompts were refined until the performance was deemed satisfactory on a validation set, including increasing the number of few-shot examples from zero to one and finally four.

The knowledge base consisted of curated oncological guideline documents for prostate and breast cancer (see Supplementary Table S5). All PDF documents were converted into plain text using the open-source library PaperMage, which works well with visually rich scientific documents. The text was chunked into overlapping segments of 512 tokens (100-token overlap) and embedded using S-PubMedBERT,[12] a transformer model pre-trained on PubMed abstracts, giving it increased semantic understanding of medical concepts, and making it highly suitable for our medical use case. Chunks were stored in a ChromaDB vector database. For evidence retrieval, cosine similarity was used to select the seven most relevant chunks for the question (in step 1) and then the seven most similar ones for each atomic fact (step 3). All LLM generations were performed using GPT-4o (gpt-4o-2024-11-20) via the OpenAI API with temperature set to 0, to ensure most deterministic outputs and reproducibility.

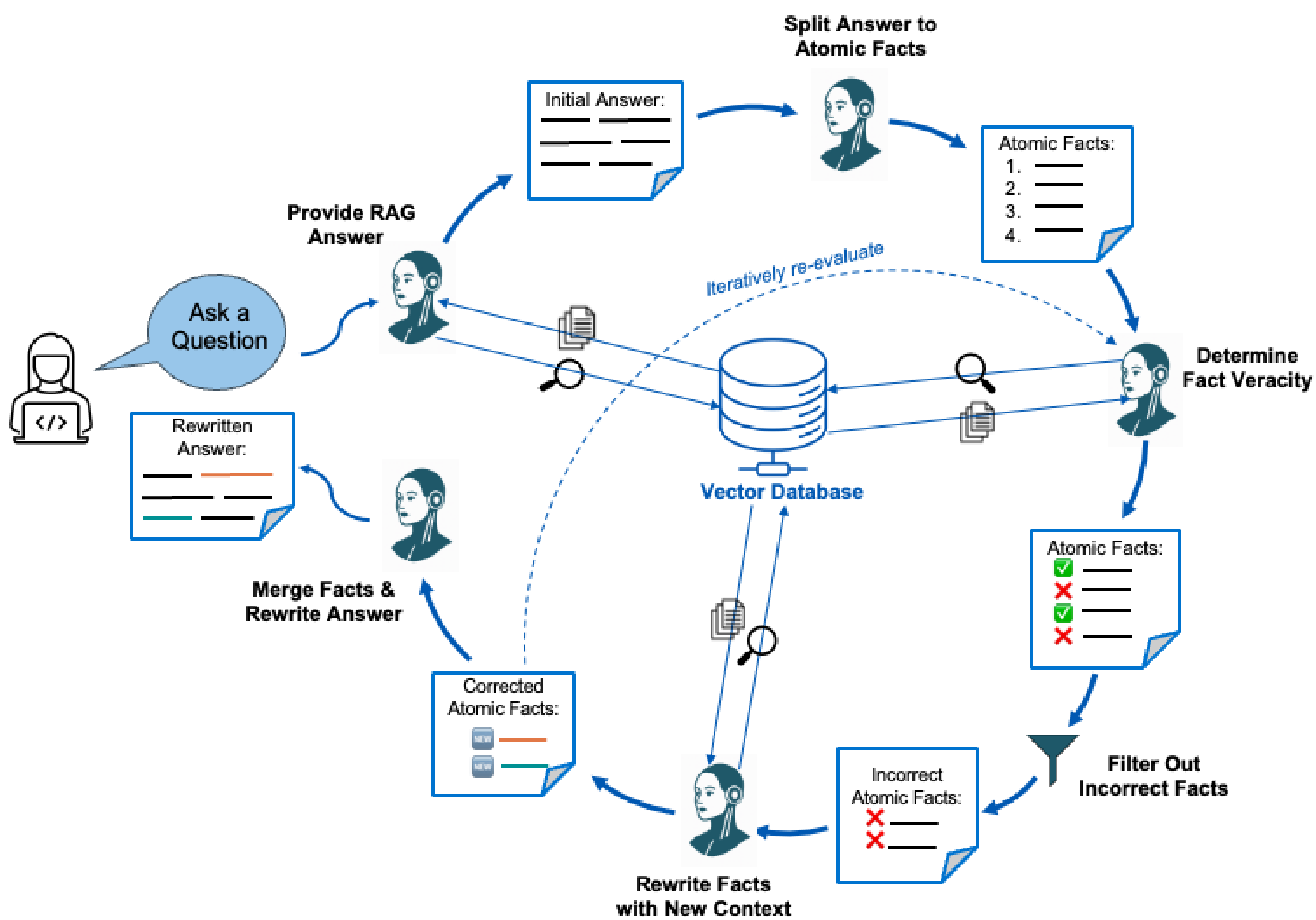

**Figure 1: Architecture diagram**
Illustration of the atomic fact checking process.

## Q&A Benchmark

The main task of this study was long-form medical question-answering. Two Q&A datasets were constructed by a medical expert (JCP), with each dataset comprising distinct questions on the diagnosis and treatment of prostate or breast cancer. Questions were based on the guideline content provided within the RAG system. They were categorized as fact-based (direct guideline knowledge) or patient-based (clinical vignettes), with varying complexity. To select the optimal prompting strategy, the first set, a validation set of 50 Q&A pairs on prostate cancer, was used. The final configuration was tested on the second set, a separate set of 60 Q&A pairs (30 prostate, 30 breast cancer). To assess the framework in a realistic clinical setting, 80 anonymized, real patient cases from a multidisciplinary tumor board were included (ethics approval 2023-626_1-S-NP). To ensure that no overfitting occurred, the pipeline was validated using a domain-unrelated set containing Neurology Q&As.[2] All Q&A sets are available in the Supplement. An overview over all used Q&As and the respective questions is available in Supplementary Table S10.

Two additional components of the pipeline were tested. Looping, where the results of one full fact-checking run (with atomic fact corrections) were used as the input to another full run, was tested to determine if it helps correct the facts that were still incorrect. Ensembling, where the fact veracity prediction (“TRUE”/”FALSE”) was based not on just one prediction output but repeated multiple predictions using the same LLM and prompt, was used to determine if fact veracity prediction could be improved.

### Human Evaluation

Fact-checking performance was evaluated by comparing assigned verdicts (“TRUE”/“FALSE”) to human assessments. Human evaluation involved verifying the correctness of each atomic verdict and categorizing errors as hallucinations, incorrect information, missing information, or a lack of context. Each atomic fact correction was analyzed for correctness, while undetected incorrect facts were considered false negatives. Final overall answers were compared to initial responses to determine whether fact-checking led to improvement, deterioration, or no change.

Validation and tumor board analyses were conducted by one medically trained scientist (AD) supervised by JCP. Test set evaluation used four independent blinded physician raters (AD, FB, JN, MN), with majority voting; disagreements (2 vs. 2) were resolved by a blinded fifth rater (JCP). All four raters agreed in 85% of cases, with ties in 4%. The Fleiss’ κ and Krippendorff’s α were both 0.16: This score is an artifact of unbalanced labels – since the LLM classifier was highly accurate, the '1' (correct) label was highly prevalent, making the “agreement expected by chance” very high.

The following metrics were computed: True Positive (TP, verdict “FALSE” confirmed by human), True Negative (TN, verdict “TRUE” confirmed by human), False Positive (FP, verdict “FALSE” not confirmed by human), and False Negative (FN, verdict “TRUE” not confirmed by human). Standard confusion matrix metrics for classification tasks (sensitivity, specificity, precision, F1, accuracy) were calculated (see Table 1, Figure 2). Annotator Guidelines are available as Supplementary Methods S1.

**Auto Evaluation**

The framework was further evaluated on the AMEGA benchmark by Fast et al.,[13] which tests adherence of LLMs to medical guidelines across 20 clinical domains. This benchmark comprises 20 patient cases (6–8 questions each, 137 in total), with 1,307 predefined evaluation criteria that can be used for automatic evaluation of generated answers. In the original study, responses were improved by iterative “question re-asking”; here, we instead applied our atomic fact-checking pipeline and compared the initial RAG answers versus corrected answers using the same criteria. The auto-evaluation was done using GPT-4o (gpt-4o-2024-11-20), and for each case, the relevant medical guidelines served as a retrieval database.

**Comparison of Open-source General vs. Medical Models**

Open-source models hold the advantages of transparency and accessibility, thus, they are relevant for on-premise deployment of AI systems in clinical settings. Hence, in addition to GPT4o, we additionally evaluated six open-source models.

Our experiments focused on differences in the fact-checking performance between medical fine-tuned open-source LLMs and their counterpart general-purpose (i.e., non-medical) models. The compared models include Gemma 3 27B vs. MedGemma 27B, Llama 3 70B vs. OpenBioLLM 70B, and Qwen 3 32B vs. Qwen 3 Medical 32B. The validation set (50 prostate Q&A pairs) was used, and a human evaluation was conducted as in previous experiments.

In addition to the quantitative fact-checking evaluation, we performed a qualitative automated evaluation. Using the LLM-as-a-judge technique, we defined seven rubrics that evaluate different aspects of the generated answers’ quality. We used GPT-4.1 as an independent judge model, which output a numeric score (range 0–1) for each answer. Each rubric used finely crafted evaluation criteria and steps, which were based on common metrics from related LLM-as-a-judge frameworks such as G-Eval, and adjusted by a medical expert (JCP) where needed. The used rubrics were: Correctness,

Completeness, Clarity, Context Faithfulness, Coherence, Medical Harmfulness, and Calibration. Evaluation steps are listed in Supplementary Table S8.

### Baseline Approach

In order to evaluate how well our atomic fact-checking framework performs compared to other methods, we compared it to the popular framework Self-Refine.[10] In this framework, the main idea is to first generate an initial response using an LLM, then use the same LLM to provide feedback on how to improve the response, and finally use the same LLM to refine the response based on that feedback. We used the prompts from the original paper, slightly adapted for the Q&A use case (prompts are in Table S4). We evaluated the approach on the AMEGA benchmark using GPT-4o, GPT-4o-mini, Gemma 3 27B, MedGemma 27B, and Llama 3.2 3B.

## Code and Data Availability Statement

The datasets and code used to perform the analyses in this study is available publicly at https://github.com/sebischair/ImprovingReliabilityMedicalQA.

## Author Contributions

JV and AD conducted the experiments and evaluation, and wrote the initial manuscript. FM and JCP provided supervision and guidance. MN, RM, JN, FB, LCA, KKB, DB, SEC, KJB contributed to data curation, data interpretation, discussed the results, and edited the manuscript.

## Acknowledgments

This study was supported with funding from the Google.org Accelerator for Generative AI (2025). The funder played no role in study design, data collection, analysis and interpretation of data, or the writing of this manuscript.

## Competing Interests

The authors declare no competing interests.